\documentclass[review]{elsarticle}

\usepackage{hyperref}
\usepackage{tabularx,ragged2e,booktabs,caption}
\usepackage{tabularx,booktabs}
\usepackage{slashbox,adjustbox}
\usepackage{multirow}
\usepackage{xcolor}
\newcolumntype{b}{X}
\newcolumntype{s}{>{{\centering\arraybackslash}\hsize=.24\hsize}X}
\newcolumntype{M}{>{{\centering\arraybackslash}\hsize=.65\hsize}X}
\newcolumntype{v}{>{\hsize=.35\hsize}X}
\newcolumntype{C}{>{\centering\arraybackslash}X} 
\makeatletter
\def\ps@pprintTitle{%
	\let\@oddhead\@empty
	\let\@evenhead\@empty
	\def\@oddfoot{\reset@font\hfil\thepage\hfil}
	\let\@evenfoot\ © 2020. This manuscript version is made available under the CC-BY-NC-ND 4.0 license. The published journal article at Computer Methods and Programs in Biomedicine journal is available from: \url{https://doi.org/10.1016/j.cmpb.2020.105725}. 
}
\makeatother

\begin{document}

\begin{frontmatter}

\title{The Effects of Skin Lesion Segmentation on the Performance of Dermatoscopic Image Classification}


\author[one]{Amirreza Mahbod\corref{mycorrespondingauthor}}
\cortext[mycorrespondingauthor]{Corresponding author}
\ead{amirreza.mahbod@meduniwien.ac.at}
\author[two]{Philipp Tschandl}
\author[three]{Georg Langs }
\author[four]{Rupert Ecker}
\author[one]{Isabella Ellinger}

\address[one]{Institute for Pathophysiology and Allergy Research, Medical University of Vienna, Vienna, Austria}
\address[two]{Department of Dermatology, Medical University of Vienna, Vienna, Austria}
\address[three]{Computational Imaging Research Lab, Department of Biomedical Imaging and Image-guided Therapy, Medical University of Vienna, Vienna, Austria}
\address[four]{Research and Development Department of TissueGnostics GmbH, Vienna, Austria}

\begin{abstract}
\noindent \textbf{Background and Objective:} Malignant melanoma (MM) is one of the deadliest types of skin cancer. Analysing dermatoscopic images plays an important role in the early detection of MM and other pigmented skin lesions. Among different computer-based methods, deep learning-based approaches and in particular convolutional neural networks have shown excellent classification and segmentation performances for dermatoscopic skin lesion images. These models can be trained end-to-end without requiring any hand-crafted features. However, the effect of using lesion segmentation information on classification performance has remained an open question. 

\noindent \textbf{Methods:} In this study, we explicitly investigated the impact of using skin lesion segmentation masks on the performance of dermatoscopic image classification. To do this, first, we developed a baseline classifier as the reference model without using any segmentation masks. Then, we used either manually or automatically created segmentation masks in both training and test phases in different scenarios and investigated the classification performances. The different scenarios included approaches that exploited the segmentation masks either for cropping of skin lesion images or removing the surrounding background or using the segmentation masks as an additional input channel for model training. 

\noindent \textbf{Results:} Evaluated on the ISIC 2017 challenge dataset which contained two binary classification tasks (i.e. MM vs. all and seborrheic keratosis (SK) vs. all) and based on the derived area under the receiver operating characteristic curve scores, we observed four main outcomes. Our results show that 1) using segmentation masks did not significantly improve the MM classification performance in any scenario, 2) in one of the scenarios (using segmentation masks for dilated cropping), SK classification performance was significantly improved, 3) removing all background information by the segmentation masks significantly degraded the overall classification performance, and 4) in case of using the appropriate scenario (using segmentation for dilated cropping), there is no significant difference of using manually or automatically created segmentation masks.   

\noindent \textbf{Conclusions:} We systematically explored the effects of using image segmentation on the performance of dermatoscopic skin lesion classification. 


\end{abstract}


\end{frontmatter}


\section{Introduction}
Skin cancer is one of the most common cancer types in the white population~\cite{leiter2014epidemiology}. Among the different types of skin cancers, malignant melanoma (MM) accounts for only a small percentage of cases, nevertheless, it is responsible for the majority of skin cancer deaths~\cite{apalla2017epidemiological}. When detected at an early stage, MM can be cured by excision of the lesion, while diagnosis at later stages is associated with a greater risk of death~\cite{leiter2014epidemiology,schadendorf2018melanoma}. Thus, early detection and accurate diagnosis of MM are crucial for the patient.

Histological examination of a skin lesion is the gold standard for diagnosis and prognosis~\cite{scolyer2019melanoma}. But, as it is an invasive, costly, and time-consuming procedure, clinicians and patients alike want to reduce the number of necessary diagnostic skin biopsies~\cite{AlAzhar2017}. The diagnostic process starts with a visual inspection of suspicious lesions by analysing the skin lesion patterns~\cite{kittler2007dermatoscopy}. 
The non-invasive and optical technique of dermatoscopy allows for a more detailed examination of the skin compared to examination by the naked eye alone and it can improve the skin lesion classification performance up to 50\%~\cite{youl2007effect}. However, even with dermatoscopes, the diagnostic performance correlates with the experience of the dermatologist~\cite{TSCHANDL2019938}. 

Automated skin lesion classification with computer-aided diagnostic (CAD) systems has been attempted in dermatology for over 30 years~\cite{Oliveira2018}. Such systems could serve as a decision aid for clinicians, particularly in enhancing the decision-making of less-experienced clinicians. Many semi or fully automatic methods have been proposed for this task~\cite{Oliveira2018}. However, proposing an accurate computerized skin lesion classification method is a challenging task due to the similar morphological appearance of different skin lesion types and also due to the various artefacts contained in dermatoscopic images. 
Originally, CAD systems for skin lesion classification were mainly based on (1) image pre-processing and artefact removal, (2) lesion segmentation, (3) feature extraction from the lesion area and (4) lesion classification using classical image processing or machine learning approaches~\cite{Oliveira2018, celebi2015state}. Various image pre-processing techniques such as colour space transformation, contrast enhancement, and image filtering were used to prepare or normalise the images for the classification~\cite{celebi2015state, CELEBI2009148}. Image border detection and segmentation were also considered as important steps for image cropping or artefact removal. Classical image processing techniques such as histogram thresholding, clustering, or active contours were widely used in the literature for skin lesion segmentation~\cite{celebi2015state}. More advanced segmentation techniques were also proposed in recent years using supervised machine learning approaches~\cite{Oliveira2018, celebi2015state, CELEBI2009148}. However, proposing an accurate skin lesion segmentation technique is a very challenging task due to several reasons such as low contrast between the lesion and its surrounding background, irregular border shapes, fuzzy borders, and fragmentation~\cite{CELEBI2009148}. After performing segmentation, skin lesion features can be extracted from the lesion area. Intensity-based features, shape-based features, and textural-based features were among the most used features for skin lesion classification. The extracted features were then used for training classifiers such as decision trees, artificial neural networks, and support vector machines~\cite{Oliveira2018, celebi2015state, CELEBI2009148, mahbod2018automatic}.

With the advent of convolutional neural networks (CNNs) and considering their excellent performance for a variety of medical classification tasks~\cite{vu2019methods, mahbod2018breast}, many CNN-based approaches have been developed to perform skin lesion classification with superior performances compared to other classical techniques~\cite{10.1001/jamadermatol.2019.1375}. In contrast to the conventional methods, many CNN-based approaches for skin lesion classification were directly applied on raw or pre-processed skin lesion images without prior image segmentation. Top performers of the International Skin Imaging Collaboration (ISIC) challenges in 2016, 2017, 2018 and 2019\footnote{\url{https://www.isic-archive.com/\#!/topWithHeader/tightContentTop/challenges} (Accessed on 2020-08-04)} are examples of such approaches.


Despite excellent classification performance of the CNN-based approaches for skin lesion classification without using any lesion segmentation masks, the potential impact of skin lesion segmentation on the performance of CNN-based classifiers has not been systematically investigated~\cite{celebi2015state}. There are only a few studies that exploited lesion segmentation information in the CNN-based classification workflow to improve the performance.

In a study by Yu {\it et al.}~\cite{7792699}, a single network was proposed that performed lesion classification in two stages. In the first stage, a very deep fully convolutional residual network was used to perform lesion segmentation. Then, the images were cropped based on the predicted segmentation masks and the cropped images were sent to a deep residual network to perform classification. The results obtained from the ISIC 2016 challenge dataset~\cite{gutman2016skin} showed improved classification when the segmentation stage was used (accuracy of 85.5\% with both stages vs. accuracy of 82.8\% with a single classification stage). However, in terms of AUC, there was only a slight improvement in the classification performance (AUC of 78.3\% with both stages vs. AUC of 78.2\% with a single classification stage). This method achieved the first rank in the ISIC 2016 challenge for the defined binary skin lesion classification task. A similar approach was proposed in~\cite{ALMASNI2020105351}. Again, two stages were used to perform lesion segmentation and the cropped images were used to perform classification. However, in this work a full resolution CNN was used as the segmentation network and other pre-trained CNNs 
were used in the classification network. Evaluated on the ISIC 2016 challenge test set and by setting the best hyper-parameters, an accuracy of 81.1\% and an AUC of 76.6\% were achieved by this method. This approach was also trained and evaluated on the subsequent ISIC challenge datasets. Applied on the ISIC 2017~\cite{Codella2017} and the ISIC 2018~\cite{codella2019skin} challenge datasets an average accuracy of 81.6\% (in comparison to 88.8\% of the ISIC 2017 challenge top performer) and 89.3\% were achieved, respectively. As the reported results for the ISIC 2018 challenge were based on the random split of the training set to training, validation, and test set, comparison to the ISIC 2018 challenge top performers is not feasible. 

Guo {\it et al.}~\cite{GUO201867} proposed a multi-channel ResNet to classify images from the ISIC 2017 challenge dataset. They performed an experiment to compare the classification results with and without a lesion detection model. The results were slightly better when the lesion detection network was used (average AUC of 87.4\% vs. 87.1\%). 
By utilising various ensembling approaches, the result was further improved to an average AUC of 91.7\%.

In the work of Chen {\it et al.}~\cite{8363769}, a multi-task framework was proposed to perform segmentation and classification within the same model. A special feature passing gate was proposed, which linked the segmentation network to the classification network to exploit useful features in the workflow. 
The method was trained and tested on the ISIC 2017 challenge dataset. Although the achieved results were superior in comparison to a single classification network (accuracy of 80.1\% vs. 77.2\%), they were inferior compared to the top performer of the ISIC 2017 challenge\footnote{The actual evaluation index of the ISIC 2017 challenge was AUC. As AUC results were not reported in~\cite{8363769}, we compared it to the top performer in terms of accuracy} (accuracy of 80.1\% vs. 88.8\%~\cite{bi2017automatic}). 

Yang {\it et al.}~\cite{8512488} proposed a multi-target CNN with three different branches to perform segmentation and two binary classification tasks for the ISIC 2017 challenge dataset. A pre-trained GoogleNet CNN was used in the encoder part of the network while the U-Net-like decoder model was used for the segmentation branches.
The reported results were superior compared to a single GoogleNet-based classification model (AUC of 88.6\% vs. 85.7\%), but inferior compared to the ISIC 2017 challenge top performer (AUC of 88.6\% vs. 91.1\% ~\cite{Matsunaga2017}). 

Díaz~\cite{8293766} proposed a network that incorporated the useful clinical information in the classification workflow. Using two segmentation networks (lesion segmentation net to produce binary lesion masks and structure segmentation net to produce eight feature segmentation masks), a very good classification result was achieved with an average AUC of 91.0\%, which ranked the approach at the 2nd place in the ISIC 2017 challenge leaderboard. However, no comparative results were reported to investigate the added value of the utilised segmentation networks for classification performance improvement.   

Burdick {\it et al.}~\cite{burdick2018rethinking} used a subset of the ISIC 2016 challenge dataset to perform a binary classification (MM vs. benign skin lesions). In their best approach, a pre-trained Inception-V3 model was fine-tuned. The training and test images were pre-processed by applying a disk morphological operation around the skin lesions of different sizes. Then, images were zero-padded and resized and finally sent to the classifier. They observed a better classification performance when the skin lesion border enlargement was applied (accuracy of 69.3\% vs. 57.3\%). However, in their study, just a small test set of 75 images was used, which could potentially lead to unreliable results. 

Tang {\it et al.}~\cite{9018274} proposed a novel Global-Part CNN to perform skin lesion classification. Their developed algorithm consisted of two sub-models which were trained sequentially. The first model (Global-CNN) was trained on resized skin lesion images using a fine-tuned Xception network. The results from this part were used for fusion and also to create class activation maps (CAMs). The created CAMs from the first model were used for probabilistic cropping of the original image to train the second model (Part-CNN). The results from these two models were fused using a weighted ensembling strategy. Applied on the ISIC 2017 challenge test set, a very good classification performance with an average AUC of 91.7\% was achieved which was further improved to 92.6\% through a data-transformed ensemble strategy.  

In the studies of Yan {\it et al.}~\cite{10.1007/978-3-030-20351-1_62} and Zhang {\it et al.}~\cite{zhang2019attention}, attention-based models were used to guide the model towards the lesion area of the image. Although no segmentation masks were directly used, the main idea of both works was to force the models to focus on the lesion area of the dermatoscopic images. Both methods were trained and evaluated on the ISIC 2017 challenge dataset and both achieved very good classification scores outperforming the top performer of the ISIC 2017 challenge. An average AUC of 91.7\% was reported in~\cite{zhang2019attention} in comparison to 91.1\% average AUC of the ISIC 2017 challenge top performer\cite{Matsunaga2017} and a MM AUC of 88.3\% was reported in~\cite{10.1007/978-3-030-20351-1_62} in comparison to 87.4\% MM AUC of the ISIC 2017 challenge top performer for MM classification~\cite{Menegola2017}. 

In contrast to studies that showed improved performances of skin lesion classification when segmentation masks were used, some other studies reported adverse effects. 
This adverse effect was specifically evident in the ISIC 2016 challenge where all top performers showed better classification performances when no lesion segmentation masks were used (c.f. by comparison of the results in sections 3 and 3B of the challenge). Li {\it et al.}~\cite{8857334} showed qualitatively that areas surrounding the skin lesions could be useful for classification so removing those areas could degrade the classification performance. Celebi {\it et al.}~\cite{CELEBI2008670} showed quantitatively that relative colour feature (the colour of a skin lesion pixel in comparison
to the average background colour) is an important feature for lesion detection and classification. Bissoto {\it et al.}~\cite{Bissoto_2019_CVPR_Workshops_org} proposed a method to perform skin lesion classification by removing the actual skin lesion from the images and just using the surrounding background information. Although their classification performance was much inferior compared to the traditional methods (i.e. using both lesion and surrounding area), it still delivered acceptable results compared to pure chance (an AUC of 77.4\% vs. an AUC of 50\% from pure chance).

In summary, the segmentation masks for skin lesion classification were either employed directly (e.g. for cropping or background removal) or indirectly (e.g. for attention-based models) to lead the networks towards the lesion area to perform the classification. However, there are some unsolved issues with the reported results. The proposed methods were either applied on a small dataset of skin lesion images or they showed inferior classification performance with a high margin in comparison to the state-of-the-art models. When the reported results showed an improvement in the classification performance, no statistical tests were performed to confirm a significant contribution of lesion segmentation to the classification performance. Therefore, the benefit of using skin lesion segmentation for improved skin lesion classification remains unsettled. 

To address these issues in this study, we first developed a baseline classification model without using any segmentation masks. We used this model as the reference and applied it on a public skin lesion dataset. Then, we utilised either perfect segmentation masks (i.e. segmentation masks created manually by the medical experts) or automatically created segmentation masks (by using one of the state-of-the-art models for skin lesion segmentation) in both training and test phases in different scenarios and investigated the classification performances. We report all classification scores as well as the significance level of the performance changes. Evaluated on the ISIC 2017 challenge test set which contained two classification tasks (i.e. MM vs. all and seborrheic keratosis (SK) vs. all), we observed several interesting outcomes. Most importantly, we observed that segmentation masks did not significantly improve the MM classification performance in any scenario while having a significant positive impact on the SK classification performance with one of the approaches where lesion segmentation masks were used for image cropping. Using the same methodical approach, we noted no significant differences in the classification performance when either manually or automatically created segmentation masks were used. Moreover, the results showed that removing all background information from the skin lesion images significantly degraded the overall classification results. 


\section{Materials and Methods}
\begin{figure*}
	\begin{center}
		\includegraphics[width=1\linewidth]{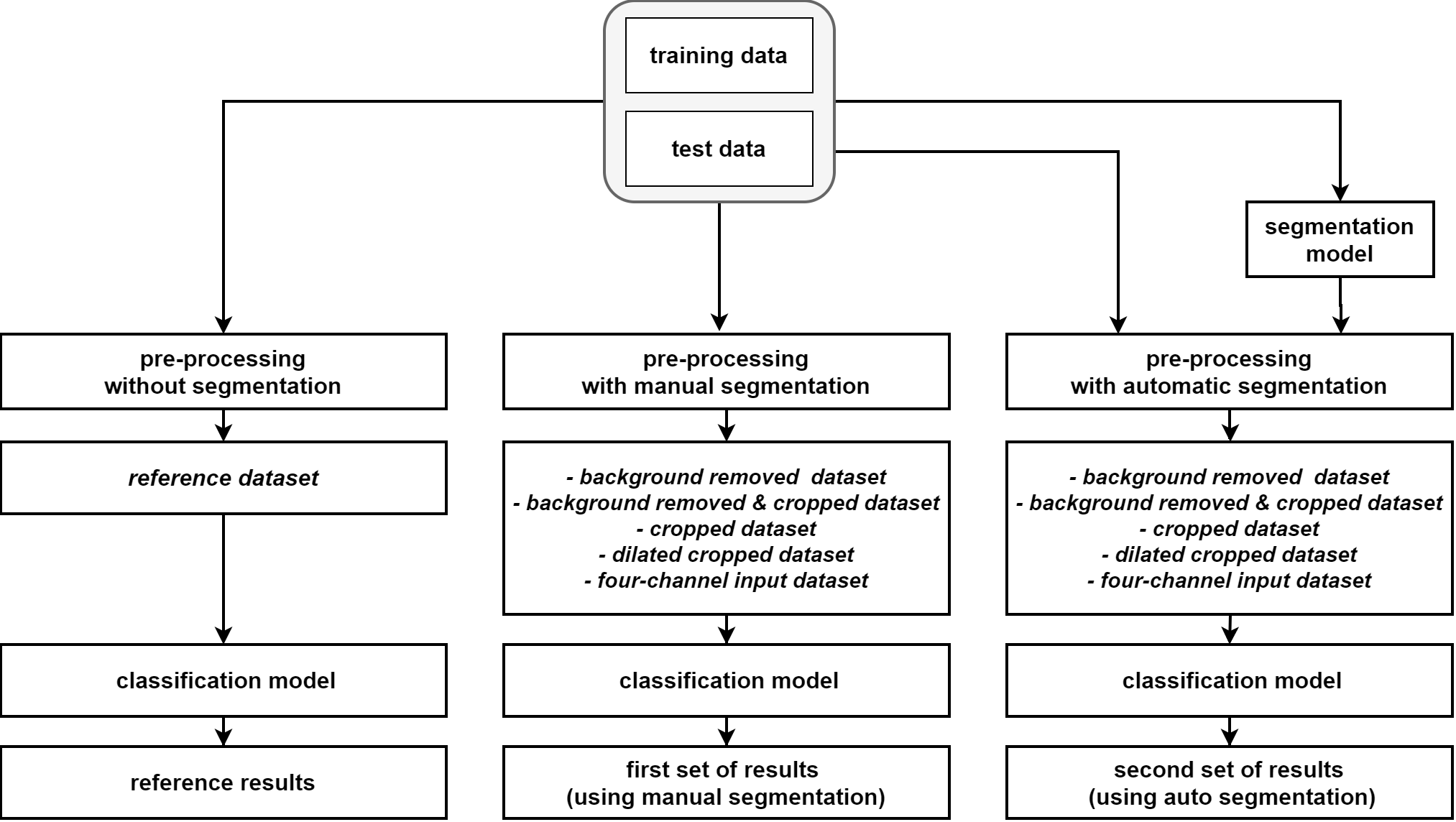}
	\end{center}
	\caption{Generic flowchart of the proposed method.}
	\label{flowchart}
\end{figure*}

The generic flowchart of the proposed method is shown in Fig.~\ref{flowchart}. Each part is described in detail in the following subsections. 
\label{material}

\subsection{Dataset}
We used a publicly available dataset to be able to compare the classification results with other state-of-the-art models that had previously been applied on the same dataset. ISIC archive is one of the biggest publicly available data sources for dermatoscopic skin lesion images. It includes the images of the well-known ISIC 2016~\cite{gutman2016skin}, ISIC 2017~\cite{Codella2017}, ISIC 2018~\cite{codella2019skin,tschandl2018ham10000,combalia2019bcn20000} and ISIC 2019 challenges~\cite{Codella2017,tschandl2018ham10000}. As we needed the labels and also the segmentation masks of both training and test sets to perform the experiments and conduct statistical tests, we chose the ISIC 2017 challenge dataset in this work.
This dataset includes most of the images from the ISIC 2016 challenge dataset as well as many additional images. The dataset comprises 2000 training images, 150 validation images, and 600 test images. We used the 2150 training and validation images in the training phase and evaluated the classification performance on the 600 test images. The ISIC 2017 challenge dataset contains three skin lesion types including MM, SK, and benign nevi (BN) classes. The 2150 images utilised in the training phase included 404 MM, 296 SK, and 1450 BN images, while the 600 test images were comprised of 117 MM, 90 SK, and 393 BN images. The images in both training and test sets contained various image artefacts 
and had different image resolutions ranging from $1022 \times 767$ to $6748 \times 4499$  pixels. 

\subsection{Pre-processing}
\label{pre-processing}

For pre-processing, we first applied the gray world colour constancy algorithm~\cite{ShervineAmidi2019,Barata2015} on all training and test images to deal with various lightening conditions in the dataset. This pre-processing step was shown to be beneficial for skin lesion classification and was used by the former ISIC challenge top performers~\cite{Matsunaga2017,gessert2018skin}. Next, we subtracted the mean RGB intensity value of the ImageNet dataset~\cite{Deng2009} from the RGB channels of all training and test images. This is a standard pre-processing technique for transfer learning~\cite{mahbod2019fusing, mahbod2019skin, mahbod2020investigating}. To create a baseline dataset without using any segmentation mask, we resized all training and test images to a fixed image size of $448 \times 448$ pixels. We used the results from this dataset as the benchmark and for comparison to all other results. We refer to this dataset as "\textit{reference dataset}" in the paper.

To investigate the effect of using image segmentation on the classification performance, we designed two sets of experiments. In the first set of experiments, we created five transformed datasets using the manual segmentation masks provided by human experts with the following details:


\begin{itemize}

	\item In the first dataset, we used the manual segmentation masks, i.e. the provided ground truth masks for the training and test sets of the ISIC 2017 challenge, to mask out the background (set all background pixels to zero) in all training and test images. Then, we resized the images to a fixed size of $448 \times 448$ pixels. We refer to this dataset as "\textit{background removed dataset}".
	
	\item In the second dataset, we used the manual segmentation masks to remove the background (set all background pixels to zero). Here, however, we used the exact lesion dimensions to crop the images. After cropping, we resized all images to $448 \times 448$ pixels. We subsequently refer to this dataset as "\textit{background removed and cropped dataset}". 
	
	\item To create the third dataset, we did not remove the background, but similar to the second dataset, we used the exact lesion dimensions to crop the images. Again, we resized all images to a fixed size of $448 \times 448$ pixels. We subsequently refer to this dataset as "\textit{cropped dataset}". 
	
	\item To create the fourth dataset, the lesions inside the segmentation masks were first dilated by a factor of 1.4 along each image dimension. Then, the dilated lesion masks were used for cropping the skin lesion images. The resulting images were then resized to $448 \times 448$ pixels. We subsequently refer to this dataset as "\textit{dilated cropped dataset}".  
	
	\item To create the fifth dataset, we used the same raw images as described for the \textit{reference dataset}. Here, we also added an additional channel to incorporate the lesion segmentation mask as the fourth channel for each individual training and test image. The images and masks were also resized to a fixed size of $448 \times 448$ pixels like in the other datasets. We subsequently refer to this dataset as "\textit{four-channel input dataset}". 
	
\end{itemize}

\begin{figure*}
	\begin{center}
		\includegraphics[width=1\linewidth]{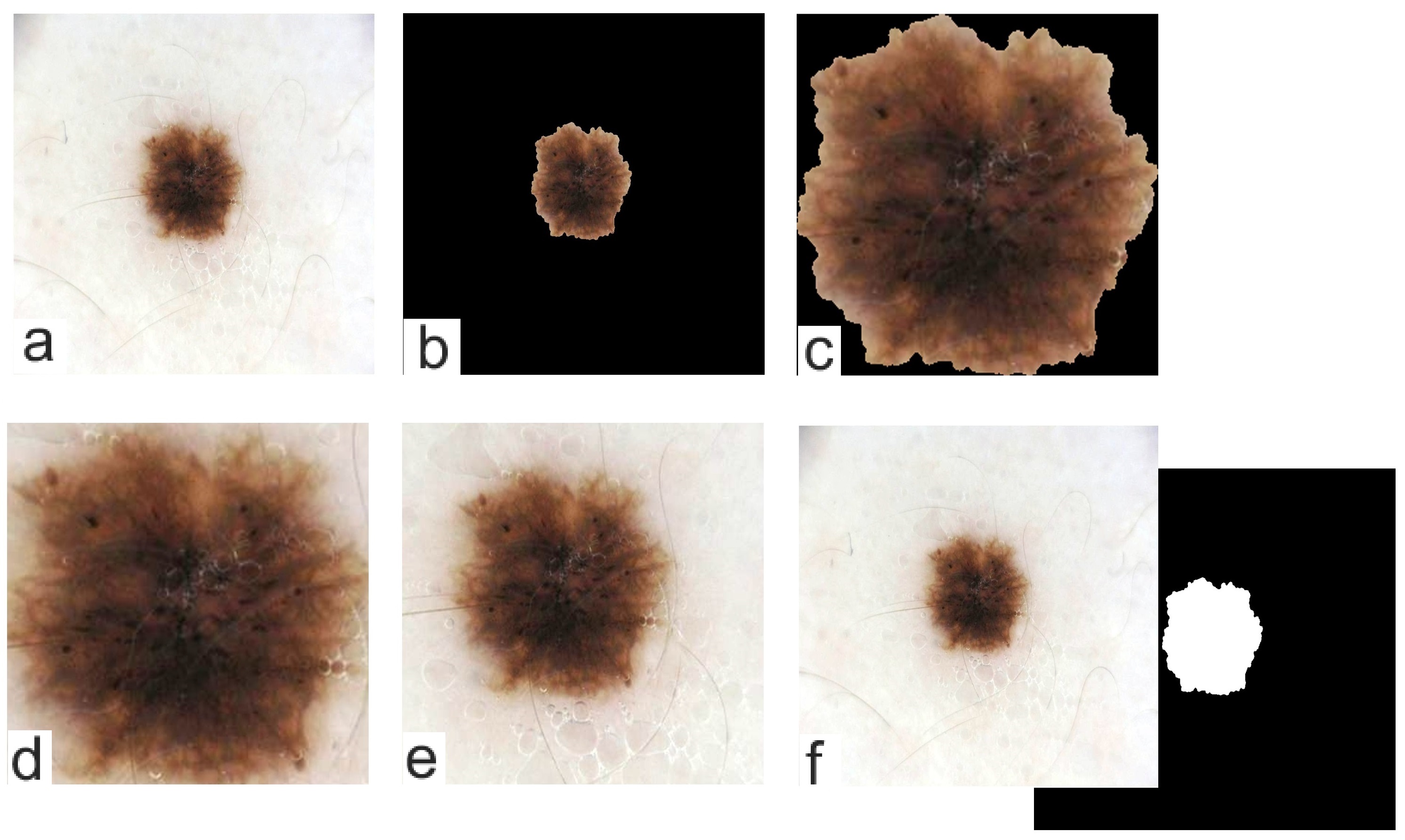}
	\end{center}
	\caption{One sample image from the reference dataset (a) and five image transformations (b-f) based on the manual segmentation mask (i.e. first experimental set). (a) raw resized image, (b) resized image with background removal, (c) resized image with background removal and lesion dilation, (d) resized image by exact lesion cropping, (e) resized image by dilated lesion cropping, (f) four-channel input data consisting of the raw resized image as well as the resized binary segmentation mask.}
	\label{images}
\end{figure*}

Fig.\ref{images} depicts an example image of the \textit{reference dataset} (Fig.\ref{images} a) and the derived image transformations (Fig.\ref{images} b-f) based on the provided manual segmentation masks.



For the second set of experiments, we created additional five transformed test datasets similar to the aforementioned datasets in the first set of experiments. However, this time instead of using the manual segmentation masks to extract the lesion information in the test images, we used one of the state-of-the-art models to perform segmentation (further details about the developed segmentation model, referred to as \textit{SkinLinkNet} can be found in Section~\ref{segmentation_model}). This step was important for two reasons. First, to investigate any difference in the classification performances by comparing the results from the perfect segmentation masks and the segmentation masks predicted by an automatic lesion segmentation model. Second, using computer-generated segmentation masks would be more suitable for the clinical practice, where the manual annotations of the lesions of the unseen dermatoscopic images are usually not provided.

\subsection{Classification model}
\label{Classification_model}
Our method developed for classification was inspired by former studies~\cite{gessert2018skin, mahbod2019fusing, MAHBOD2020105475} which had shown excellent classification performances. In this work, we avoided using any external dataset or sophisticated ensembling strategies as we did not aim to achieve the best classification performance. Instead, we developed a baseline single classification model to be used as a reference model in all experiments. For the pre-trained model selection, we used the shallowest version of the EfficientNet family (EfficientNetB0)~\cite{tan2019efficientnet} and fine-tuned it with the training set. For the fine-tuning, we removed the FC layer of the pre-trained network and then used a global average pooling layer to connect it to two blocks of batch normalisation layers, dropout layers, and FC layers. We used a drop out factor of 0.3 in both dropout layers. 64 and 3 neurons were used in the first and second FC layers, respectively. For the \textit{four-channel input dataset}, where four-channel input images (RGB channels plus the mask channel) were used, we added a $3 \times 3 \times 3$ convolutional layer to convert the 4 channel inputs to 3 channel data and then connected it to the utilised EfficientNetB0 model. We initialised the weights of the newly added layers by Xavier initialisation method~\cite{glorot2010understanding} and kept the learning rate of these newly added layers 10 times larger compared to all other learnable layers. We used Adam optimisation method~\cite{Kingma2014} and trained the networks for 70 epochs. To deal with the unbalanced training data, we used weighted focal loss function~\cite{lin2017focal}. We used five-fold cross-validation and for each fold, we saved the best model by monitoring the average AUC score for the validation set. In the inference phase, first, we applied 50-fold test time augmentation and then, the augmented test images were sent to the saved five sub-models (one model for each fold) and the average results were used as the final prediction vectors.  The utilised augmentation techniques in training and test phases included both morphological and colour augmentations such as random scaling (scale limit of 0.1 with the probability of 0.3), random rotations (0, 90, 180, 270 degrees with the probability of 0.5), vertical and horizontal flipping (with the probability of 0.5), brightness and contrast shifts (brightness and contrast limit of 0.15 with the probability of 0.4), and random adaptive histogram equalisation (tile grid size of 8 with the probability of 0.1). Further details about the utilised approach for fine-tuning and test time augmentation can be found in~\cite{MAHBOD2020105475}. This developed classification model was used in all experiments in both training and test phases.

\subsection{Segmentation masks}
\label{segmentation_model}
We used two types of segmentation masks to create the datasets as described in Section~\ref{pre-processing} (i.e. manually and automatically created segmentation masks). 

For the first set of experiments, we used the provided ground truth segmentation masks of the ISIC 2017 challenge for both training and test images. Although using manually annotated segmentation masks in the test phase is not a practical approach in the clinical setting, it enabled us to reveal the potential impact of the perfect segmentation masks on the classification performance.

In the second set of experiments, we used one of the state-of-the-art skin lesion segmentation models. We developed an identical approach as explained in~\cite{TSCHANDL2019111}. We used the LinkNet-152 segmentation model~\cite{8305148} and pre-trained the encoder part of the model with the HAM10000 dataset~\cite{tschandl2018ham10000}. For training the full segmentation model, we used the 2000 training images of the ISIC 2017 challenge and monitored the segmentation performance on the 150 images of the validation set of the ISIC 2017 challenge dataset. We used identical hyper-parameters, pre-processing, and data augmentation for training as described in~\cite{TSCHANDL2019111}. We resized all images and their corresponding masks to $512 \times 512$ pixels using bilinear interpolation as a pre-processing step. For data augmentation, we used horizontal and vertical flipping as well as random 90-degrees rotations. A combination of binary cross-entropy and Jaccard loss was used for model training. We applied the trained model to perform segmentation of the 600 test images of the ISIC 2017 challenge. We used the predicted segmentation masks to create the datasets (as explained in Section~\ref{pre-processing}) for the second set of experiments. We refer to this model as \textit{SkinLinkNet} in the rest of the paper. For reporting the results for the automatic segmentation model in the results section, we mainly used \textit{SkinLinkNet}. 

To extend our comparison, we also developed two other segmentation models, namely \textit{SkinUNet} and \textit{SkinFPN+} and compared their performance with the \textit{SkinLinkNet} model. 

For \textit{SkinUNet}, we developed a modified U-Net model~\cite{Ronneberger2015}. In the encoder part of the model, a pre-trained ResNet34~\cite{He2016} network was used. We used Adam optimiser and a combination of dice loss and focal loss to train the model. Similar to \textit{SkinLinkNet}, we used the resized $512 \times 512$ pixel images from the ISIC 2017 dataset to train the model. As augmentation, horizontal and vertical flipping (with the probability of 0.5), random brightness and contrast shift (brightness and contrast limit of 0.15 with the probability of 0.4), random 0-, 90-, 180-, and 270-degree rotations (with the probability of 0.5), and random adaptive histogram equalization (tile grid size of 8 with the probability of 0.1) were used. 

For the \textit{SkinFPN+}, we used the training scheme similar to the \textit{SkinUNet} with two main differences. First, we used the feature pyramid network (FPN)~\cite{8099589, kirillov2017unified} as the main architecture with the pre-trained ResNet34 network in the encoder part of the model. Second, we used extensive external data to train this segmentation model (Hence we used the "+" sign in the model's name). Besides the ISIC 2017 training and validation images, we used the recently released HAM10000 segmentation masks and images to train this network~\cite{tschandl2020human}. Similar to the ISIC 2017 dataset, all 10015 images and segmentation masks of the HAM10000 dataset were resized to $512 \times 512$ pixel as a pre-processing step. 

We developed these two additional segmentation models to have slightly superior and slightly inferior segmentation models in comparison to the \textit{SkinLinkNet} model (refer to Table \ref{segmentation_ref} for quantitative comparison). It is worth mentioning that in the first set of experiments, we used perfect manual segmentation masks that can be considered as the best possible segmentation model. 

\subsection{Evaluation}
For evaluating the classification performances, we used AUC as the main evaluation index identical to the ISIC 2017 challenge. Although we trained all models to solve a ternary classification task, as suggested in the ISIC 2017 challenge, we calculated the AUCs for two binary classification problems (i.e. MM vs. all and SK vs. all). To convert the ternary classification vectors to two binary classification vectors, we used a one-versus-all approach. In addition to AUC, we also calculated accuracy, sensitivity, and specificity as additional evaluation indexes. To measure those indexes, we converted the classification probability vectors to binarize classification vectors by using a threshold of 0.5. For comparing the statistical differences between different AUCs, we employed the method described by Delong {\it et al.}~\cite{delong1988comparing} and to measure the statistical significance levels of the other evaluation indexes, we used McNemar statistical test~\cite{fagerland2013mcnemar}. Delong {\it et al.} non-parametric approach is widely used for comparing empirical ROC curves and AUC scores of paired samples. This method became popular as it does not have the normality assumption for the sample distribution among the two classes. The statistical significance level (p-value) of the Delong {\it et al.} method can be derived from the following formulas: 

\begin{equation}
z = \frac{A_1 - A_2}{\sqrt{V(A_1) + V(A_2) -2cov (A_1, A_2)}}
\end{equation}

where $A_1$ and $A_2$ are the empirical AUCs of the first and second tests and $V$ and $Cov$ are the variance and covariance functions. The p-value is then calculated as $2 (1 - \phi(|z|))$ where $\phi$ is the standard normal cumulative distribution function~\cite{delong1988comparing, demler2012misuse}. McNemar is a method that statistically compares predicted labels against ground truth labels for binary matched-pairs data then detects whether the misclassification rates between the two tests are statistically significant or not~\cite{fagerland2013mcnemar}. As this method can be used for binary matched-pair samples, it is a good choice for comparing the accuracy, sensitivity, and specificity of two models. 

For measuring the segmentation performances, we calculated the average dice score and average Jaccard score as the main evaluation indexes identical to the ISIC 2017 challenge. To perform statistical tests on the segmentation results, we used Wilcoxon signed-rank test method~\cite{gibbons2014nonparametric}. Wilcoxon signed-rank statistical test method is a non-parametric test for two populations when samples are paired (in this study we have a dice score and Jaccard score for each sample). The test statistic in the method is the sum of the ranks of positive differences between the Jaccard score and the dice score in the two populations. Further details about this method can be found in~\cite{gibbons2014nonparametric}. In all statistical tests, a two-sided significance level of 5\% (p-value = 0.05) was used as the threshold.

\subsection{Implementation}
Keras\footnote{\url{https://keras.io/} (Accessed on 2020-08-04)} deep learning framework was used in the development of the classification model as well as the development of the \textit{SkinUNet} and \textit{SkinFPN+} segmentation models. For developing the \textit{SkinLinkNet} segmentation model, PyTorch framework was used. All pre-processing steps were performed offline and with MATLAB software (version 2018a). We used MedCalc (version 19.1) and MATLAB software (version 2018a) to perform the statistical tests. All experiments were conducted on a single workstation with an Intel Core i7-8700 3.20 GHz CPU, 32 GB of RAM, and a TITIAN V NVIDIA GPU card with 12 GB of installed memory. Training and test time for each experiment were $\approx110$ minutes and $\approx6.5$ minutes, respectively. 

\section{Results}
\label{results}
All reported results in this section are based on the 600 images of the ISIC 2017 challenge test set. The \textit{SkinLinkNet} segmentation model was used to report the results for the cases where automatically created segmentation masks were used unless stated otherwise in the text.

We started the experiments by evaluating the classification and segmentation performances of the utilised models. 

As mentioned in Section~\ref{material}, we did not aim to improve the classification results compared to the existing models as it was not the primary aim of this study. As the consequence, we did not use any external datasets or sophisticated ensembling strategy to improve the classification performance. On the other hand, we did not want to have a model that delivers much inferior performance in comparison to the other state-of-the-art algorithms. We developed a baseline classification model that produced results comparable to the state-of-the-art models. Table~\ref{classification_ref} shows the comparison between the classification performance of the utilised model and the top three performers of the ISIC 2017 classification challenge (rows 1--3)\cite{Matsunaga2017, Menegola2017, Diaz2017} as well as four other methods (rows 4--7)~\cite{ 8293766, 9018274, 10.1007/978-3-030-20351-1_62, mahbod2019fusing}. The aforementioned four methods had been developed after the competition and all had shown better overall classification performances in comparison to the ISIC 2017 challenge top-ranked team. The last row in Table~\ref{classification_ref} shows our baseline classifier performance that was trained with the \textit{reference dataset}. 

Table~\ref{segmentation_ref} compares the results from the main utilised segmentation algorithm (i.e. \textit{SkinLinkNet} in the second set of experiments) with the top three performers of the ISIC 2017 segmentation challenge~\cite{bi2017automatic, yuan2017automatic, berseth2017isic} as well as the two additional developed segmentation models explained in Section~\ref{segmentation_model}. The significance levels (p-value) in Table~\ref{segmentation_ref} show whether results obtained with state-of-the-art algorithms are significantly different from the \textit{SkinLinkNet} model.


\begin{table*}[]
	\caption{Performance comparison of the baseline classification model used in this work (Mahbod {\it et al.}) (last row) with the top three performers of the ISIC 2017 classification challenge (row 1-3) as well as four other state-of-the-art algorithms (row 4-7) based on the area under the receiver operating characteristic curve (AUC) scores for malignant melanoma (MM) vs. all classification task and seborrheic keratosis (SK) vs. all classification task. The reported p-values for the classification tasks were calculated with Delong {\it et al.}~\cite{delong1988comparing} method by comparing the AUC of the utilised baseline classification model (Mahbod {\it et al.} (last row)) with the AUC of other approaches (pair-wise comparison). As classification prediction vectors of the first and third ranks of the challenge were not available, statistical comparison with those two approaches was not possible.}
	\label{classification_ref}
	\begin{tabular}{lccccc}
		\hline
		\multicolumn{1}{c}{\textbf{ \small Method}} & \textbf{\begin{tabular}[c]{@{}c@{}} \small MM AUC\\  \small (\%)\end{tabular}} & \textbf{\begin{tabular}[c]{@{}c@{}} \small P-value \\ \small (MM)\end{tabular}} & \textbf{\begin{tabular}[c]{@{}c@{}} \small SK AUC \\  \small (\%)\end{tabular}} & \textbf{\begin{tabular}[c]{@{}c@{}} \small P-value \\  \small (SK)\end{tabular}} & \textbf{\begin{tabular}[c]{@{}c@{}} \small Avg. AUC\\  \small (\%)\end{tabular}} \\ \hline
		 \small Matsunga {\it et al.}~\cite{Matsunaga2017}          &  86.8                                                       &  n/a                                                        &  95.3                                                       &  n/a                                                        &  91.1                                                         \\ 
		 \small Gonzales {\it et al.}~\cite{Diaz2017}                      &  85.6                                                       &  0.33                                                       &  \textbf{96.5}                                              &  0.17                                                       &  91.0                                                         \\ 
		 \small Menegola {\it et al.}~\cite{Menegola2017}                      &  87.4                                              &  n/a                                                        &  94.3                                                       &  n/a                                                        &  90.8                                                         \\ \hline 
		 \small Mahbod {\it et al.}~\cite{mahbod2019fusing}                               &  87.3                                                       & 0.97                                                          &  95.5                                                       & 0.63                                                          &  91.4                                                \\ 
		 \small Gonzales {\it et al.}~\cite{8293766}                               &  87.3                                                       & 0.97                                                          &  96.2                                                       & 0.27                                                          &  91.7                                                \\  
		 \small Yan {\it et al.}~\cite{10.1007/978-3-030-20351-1_62}                            &  88.3                                                     & 0.47                                                          &  n/a                                                       & n/a                                                          &  n/a                                               \\  
		 \small Tang {\it et al.}~\cite{9018274}                            &  \textbf{88.9}                                                       & 0.28                                                          &  96.4                                                       & 0.14                                                          &  \textbf{92.6}                                               \\ \hline 
		 \small Mahbod {\it et al.} (This study)                               &  87.2                                                       & --                                                          &  95.1                                                       & --                                                          &  91.2                                               \\ \hline
	\end{tabular}
\end{table*}

\begin{table}[]
	\caption{Performance comparison of the main segmentation model used in this work, \textit{SkinLinkNet} (last row), with the top three performers of the ISIC 2017 segmentation challenge (row 1-3) as well as two additional developed segmentation models (row 4-5) based on the average Dice and average Jaccard index. The reported p-values were calculated with Wilcoxon signed-rank test method~\cite{gibbons2014nonparametric} by pair-wise comparison of the Jaccard and Dice scores of the \textit{SkinLinkNet} model with the other approaches. Information regarding the Jaccard and Dice scores of the top three performers were derived from the ISIC 2017 challenge leaderboard.  For those cases where the results were significantly inferior compared to the \textit{SkinLinkNet} results, the p-values are shown in red and for those cases where the results were significantly superior compared to the \textit{SkinLinkNet} results, the p-values are shown in blue.}
	\label{segmentation_ref}
	\begin{tabular}{lcccc}
		\hline
		\multicolumn{1}{c}{\small \textbf{ Method}} & \textbf{\begin{tabular}[c]{@{}c@{}} \small Avg. Jaccard\\ \small  (\%)\end{tabular}} & \textbf{\begin{tabular}[c]{@{}c@{}} \small P-value\\  \small (Jaccard)\end{tabular}} & \textbf{\begin{tabular}[c]{@{}c@{}}\ \small Avg. Dice\\  \small (\%)\end{tabular}} & \textbf{\begin{tabular}[c]{@{}c@{}} \small P-value \\  \small (Dice)\end{tabular}} \\ \hline
		 \small Yading Yuan~\cite{yuan2017automatic}&  76.5 $\pm$ 19.6                                                  &  0.052                                                        &  84.9 $\pm$ 16.6                                                &  0.052                                                        \\ 
	 \small Matt Berseth~\cite{berseth2017isic}& 76.2 $\pm$  19.7                                                         &  0.21                                                         &  84.7 $\pm$  16.4                                                      &  0.27                                                         \\ 
		 \small Bi {\it et al.}~\cite{bi2017automatic}& 76.0  $\pm$  20.6                                                        &   0.99                                                         &  84.4 $\pm$ 17.4                                                       &   0.79                                                         \\ \hline
		
		 
		  \small \textit{SkinUNet}  &  73.3 $\pm$ 23.1       & {\color{red}0.022}    &  81.9 $\pm$ 21.0      & {\color{red}0.025}    \\ 
		  
		  \small \textit{SkinFPN+}  &  \textbf{77.3 $\pm$ 18.8}       & {\color{teal} $\textless$0.001}    &  \textbf{85.6 $\pm$ 15.5}      & {\color{teal}$\textless$0.001}    \\ \hline
		 
		  \small\textit{SkinLinkNet}                             &  76.0 $\pm$ 19.5                                                           & --            &  84.5 $\pm$ 16.5                                                        & --                                                            \\ \hline
	\end{tabular}
\end{table}

Results displayed in Table~\ref{auc_compare} show the main findings of this study. The classification performances of the different approaches are compared based on the AUC scores. All reported p-values were derived from comparisons of the results obtained from the \textit{reference dataset} with the other approaches using Delong {\it et al.} method~\cite{delong1988comparing}. The comparison of the results based on the evaluation indexes accuracy, sensitivity and specificity is displayed in Table~\ref{acc_compare}. For these results, the reported significance levels were calculated by comparing the results from the \textit{reference dataset} with the other approaches using McNemar statistical test method~\cite{fagerland2013mcnemar}.

\begin{table*}[]
	\caption{Comparison of the classification performances based on the area under the receiver operating characteristic curve (AUC) scores. The results in the first row show the classification performance without using any segmentation masks (i.e using \textit{reference dataset} for both training and test phases). The results in row 2--6 show the classification performances of the first set of experiments where manual segmentation masks were used in both training and test phases. The results in row 7--11 show the classification performances of the second set of experiments where the \textit{SkinLinkNet} model was utilised (details in section~\ref{segmentation_model}). The best classification scores for each task are shown in bold. The reported p-values were derived by comparison of the results from the \textit{reference dataset} with the other approaches using Delong {\it et al.} method~\cite{delong1988comparing}. For those cases where the results were significantly inferior compared to the reference results, the p-values are shown in red and for those cases where the results were significantly superior compared to the reference results, the p-values are shown in blue.  \textbf{Abbreviations} ref: \textit{reference dataset}; bg rm: \textit{background removed dataset}; bg rm \& crop: \textit{background removed and cropped dataset}; crop: \textit{cropped dataset}; dilated crop: \textit{dilated cropped dataset}; 4-channel: \textit{four-channel input dataset}; MM: malignant melanoma; SK: seborrheic keratosis; AUC: area under the receiver characteristic operating curve.}
	\label{auc_compare}
	\begin{tabular}{lcccccl}		
		\hline \hline
		\multicolumn{1}{c}{\textbf{Dataset}}   & \textbf{\begin{tabular}[c]{@{}c@{}}Segmentation\\ Model\end{tabular}} & \textbf{\begin{tabular}[c]{@{}c@{}}MM AUC\\  (\%)\end{tabular}} & \textbf{\begin{tabular}[c]{@{}c@{}} P-value\\  (MM)\end{tabular}} & \textbf{\begin{tabular}[c]{@{}c@{}}SK AUC\\  (\%)\end{tabular}} & \textbf{\begin{tabular}[c]{@{}c@{}} P-value\\  (SK)\end{tabular}} & \multicolumn{1}{c}{\textbf{\begin{tabular}[c]{@{}c@{}}Avg. AUC \\ (\%)\end{tabular}}} \\ \hline \hline
		ref             & None     & 87.2          &  --               & 95.1         &  --                 & 91.2      \\ \hline 
		bg rm           & Manual   & 82.5          &{\color{red}0.02}  & 93.0         & 0.11                &  87.7       \\ 
		bg rm \& crop   & Manual   & 87.5          &  0.89             & 93.6         & 0.24                & 90.5         \\ 
		crop            & Manual   & 86.4          &  0.64             & 95.8         & 0.44                & 91.1          \\ 
		dilated crop     & Manual   & \textbf{89.4} &  0.09             & \textbf{96.7}& {\color{teal}0.02} & \textbf{93.0}  \\ 
		4-channel       & Manual   & 87.2          &  0.98             & 92.9         & {\color{red}0.01}   & 90.0            \\ \hline 
		bg rm           & \textit{SkinLinkNet}     & 79.0          &{\color{red}0.0003}& 93.4         & 0.24                & 86.2             \\ 
		bg rm \& crop   & \textit{SkinLinkNet}     & 85.0          &  0.25             & 95.1         & 0.99                & 90.1              \\  
		crop            & \textit{SkinLinkNet}     & 87.1          &  0.94             & 96.1         &  0.22               & 91.6               \\  
		dilated crop     & \textit{SkinLinkNet}     & \textbf{88.7} &   0.22            &\textbf{96.6} & {\color{teal}0.03} & \textbf{92.6}       \\ 
		4-channel       & \textit{SkinLinkNet}     & 81.2          &{\color{red}0.0008}& 94.7         & 0.64                &   87.8               \\ \hline \hline 
	\end{tabular}
\end{table*}

\begin{table*}[]
	\caption{Comparison of the classification performances based on the accuracy, sensitivity, and specificity scores. The results in the first row show the classification performance without using any segmentation masks (i.e using \textit{reference dataset} for both training and test phases). The results in row 2--6 show the classification performances of the first set of experiments where manual segmentation masks were used in both training and test phases. The results in row 7--11 show the classification performances of the second set of experiments where the \textit{SkinLinkNet} model was utilised (details in section~\ref{segmentation_model}). The best classification scores for each task are shown in bold. The reported p-values were derived from comparing the results from the \textit{reference dataset} with the other approaches using McNemar statistical test~\cite{fagerland2013mcnemar}. Results that were significantly better than the reference results are shown in blue.  \textbf{Abbreviations} ref: \textit{reference dataset}; bg rm: \textit{background removed dataset}; bg rm \& crop: \textit{background removed and cropped dataset}; crop: \textit{cropped dataset}; dilated crop: \textit{dilated cropped dataset}; 4-channel: \textit{four-channel input dataset}; MM: malignant melanoma; SK: seborrheic keratosis; Acc: accuracy (\%); Sen: sensitivity (\%); Spec: specificity (\%); Seg model: segmentation model.}
	\label{acc_compare}
	\begin{tabular}{lccccccccc}
		\hline
		\hline
		\multicolumn{1}{c}{\textbf{Dataset}}   & \textbf{\begin{tabular}[c]{@{}c@{}}Seg\\ model\end{tabular}} & \textbf{\begin{tabular}[c]{@{}c@{}}MM \\ Acc\end{tabular}} & \textbf{\begin{tabular}[c]{@{}c@{}}SK \\ Acc\end{tabular}} & \textbf{\begin{tabular}[c]{@{}c@{}}MM \\  Sen\end{tabular}} & \textbf{\begin{tabular}[c]{@{}c@{}}SK \\  Sen\end{tabular}} & \textbf{\begin{tabular}[c]{@{}c@{}}MM \\ Spec\end{tabular}} & \textbf{\begin{tabular}[c]{@{}c@{}}SK \\ Spec\end{tabular}} & \textbf{\begin{tabular}[c]{@{}c@{}}P-value\\ (MM)\end{tabular}} & \multicolumn{1}{l}{\textbf{\begin{tabular}[c]{@{}l@{}}P-value\\ (SK)\end{tabular}}} \\ \hline \hline
		ref           & None    & 85.7         & 91.0          & 67.5          & 74.4          & 90.1         & 93.2          & --   & --  \\ \hline
		bg rm         & Manual  & 83.9         & 90.2          & 59.0          & 63.3          & 89.9         & 94.9          & 0.19 & 0.45     \\
		bg rm \& crop & Manual  & 86.7         & 91.3          & 70.9          & 64.4          & 90.5         & 96.1          & 0.50 & 0.78      \\
		crop          & Manual  & 84.8         &\textbf{92.0}  &\textbf{75.2}  &\textbf{81.1}  & 87.2         & 93.9          & 0.55 & 0.39        \\
		dilated crop   & Manual  &\textbf{87.7} &\textbf{92.0}  & 73.5          & 78.9          & \textbf{91.1}& 94.3          & 0.09 & 0.32        \\
		4-channel     & Manual  & 84.7         & 91.2          & 58.1          & 61.1          & \textbf{91.1}& \textbf{96.5} & 0.56 & 0.39     \\ \hline
		bg rm         & \textit{SkinLinkNet}    & 84.3         & 90.7          & 57.2          & 63.3          & 90.9         & 95.5          & 0.56 & 0.39           \\
		bg rm \& crop & \textit{SkinLinkNet}    & 85.7         & 92.0          & 57.3          & 56.7          & 92.5         & \textbf{98.2} & 0.99 & 0.40                     \\
		crop         & \textit{SkinLinkNet}    & 84.7         & 92.3          & 67.5          & 77.8          & 88.8         & 94.9          & 0.45 & 0.21        \\
		dilated crop   & \textit{SkinLinkNet}    &\textbf{85.8} & \textbf{93.3} & \textbf{66.7} & 77.8          & 90.5         & 96.1          & 0.88 & {\color{teal}0.02}      \\
		4-channel     & \textit{SkinLinkNet}    & 84.3         & 89.1          & 36.7          &\textbf{78.9}  &\textbf{95.9} & 91.0          & 0.40 & 0.11  \\ \hline \hline
	\end{tabular}
\end{table*} 

As apparent from Table~\ref{auc_compare}, the best overall classification performance was achieved when the \textit{dilated cropped dataset} was used.  Using \textit{dilated cropped dataset}, the results also showed slightly superior performance when manual segmentation masks were used in comparison with automatically created segmentation masks by \textit{SkinLinkNet} (an average AUC of 93.0\% vs. 92.6\%). To have more extensive comparison for dilated cropping, we also performed additional experiments with the other two developed segmentation models (i.e. \textit{SkinUNet} and \textit{SkinFPN+}) and report the results in Table \ref{dilated_crop}.  As the results from various approaches with the \textit{dilated cropped dataset} were very competitive, we performed statistical tests to investigate the significance level of the performance differences. The results of these tests are depicted in Table~\ref{manualVSauto}.

\begin{table*}[]
	\caption{Comparison of the classification performances based on the area under the receiver operating characteristic curve (AUC) scores. The results in the first row show the classification performance without using any segmentation masks (i.e. using \textit{reference dataset} for both training and test phases). The results in row 2--5 show the classification performances when \textit{dilated cropped dataset} was used. The reported p-values were derived by comparison of the results from the \textit{reference dataset} with the other approaches using Delong {\it et al.} method~\cite{delong1988comparing}. Results that were significantly better than the reference results are shown in blue. \textbf{Abbreviations} ref: \textit{reference dataset}; dilated crop: \textit{dilated cropped dataset}; MM: malignant melanoma; SK: seborrheic keratosis; AUC: area under the receiver characteristic operating curve.}
	\label{dilated_crop}
	\begin{tabular}{lcccccl}		
		\hline \hline
		\multicolumn{1}{c}{\textbf{Dataset}}     & \textbf{\begin{tabular}[c]{@{}c@{}}MM AUC\\  (\%)\end{tabular}} & \textbf{\begin{tabular}[c]{@{}c@{}} P-value\\  (MM)\end{tabular}} & \textbf{\begin{tabular}[c]{@{}c@{}}SK AUC\\  (\%)\end{tabular}} & \textbf{\begin{tabular}[c]{@{}c@{}} P-value\\  (SK)\end{tabular}} & \multicolumn{1}{c}{\textbf{\begin{tabular}[c]{@{}c@{}}Avg. AUC \\ (\%)\end{tabular}}} \\ \hline \hline
		ref (no Segmentation model)                   & 87.2          &  --               & 95.1         &  --                 & 91.2      \\ \hline 
	
		dilated crop (by manual masks)        & 89.4 &  0.09             & 96.7& {\color{teal}0.02} & 93.0 \\ 
	  
		dilated crop (by \textit{SkinLinkNet})         & 88.7 &   0.22            &96.6 & {\color{teal}0.03} & 92.6       \\ 
		
		dilated crop (by \textit{SkinUNet})          &88.3  & 0.16              &96.6 &{\color{teal}0.02}  &92.7        \\ 
		
		dilated crop (by \textit{SkinFPN+})         &89.2  &  0.08             &96.9 &{\color{teal}0.01}  & 93.0       \\ 
	 \hline \hline 
	\end{tabular}
\end{table*}

\begin{table}[]
	\caption{Comparison of the classification results for \textit{dilated cropped dataset} using either manually created segmentation masks or automatically created segmentation masks. The reported p-values were derived by comparison of the results from the manual segmentation masks with the other approaches using Delong {\it et al.} method~\cite{delong1988comparing} \textbf{Abbreviations} dilated crop: \textit{dilated cropped dataset}; MM: malignant melanoma; SK: seborrheic keratosis; Acc: accuracy (\%); AUC: area under the receiver characteristic operating curve (\%); Seg model: segmentation model.}
	\label{manualVSauto}
	\begin{tabular}{ccccccc}
		\hline
		\textbf{\begin{tabular}[c]{@{}c@{}}Seg model\\ (for dilated crop)\end{tabular}} & \textbf{\begin{tabular}[c]{@{}c@{}}Avg. \\ AUC\end{tabular}} & \textbf{\begin{tabular}[c]{@{}c@{}}Avg. \\ Acc\end{tabular}} & \textbf{\begin{tabular}[c]{@{}c@{}}P-value\\  (MM AUC)\end{tabular}} & \textbf{\begin{tabular}[c]{@{}c@{}}P-value\\  (SK AUC)\end{tabular}} & \textbf{\begin{tabular}[c]{@{}c@{}}P-value\\  (MM Acc)\end{tabular}} & \textbf{\begin{tabular}[c]{@{}c@{}}P-value\\  (SK Acc)\end{tabular}} \\ \hline
		Manual              & 93.0  & 89.9   & --       & --     & --        & --        \\
		\textit{SkinLinkNet} & 92.6  & 89.6   & 0.29     & 0.83   & 0.06      & 0.11   \\
		\textit{SkinUNet}    & 92.7  & 89.8   & 0.36     & 0.85   & 0.84      & 0.81  \\
		\textit{SkinFPN+}    & 93.0  &  89.8  & 0.73     & 0.53   & 0.66      & 0.55    \\ \hline
	\end{tabular}
\end{table}

\begin{figure}[t]
	\begin{center}
		\includegraphics[width=0.8\linewidth]{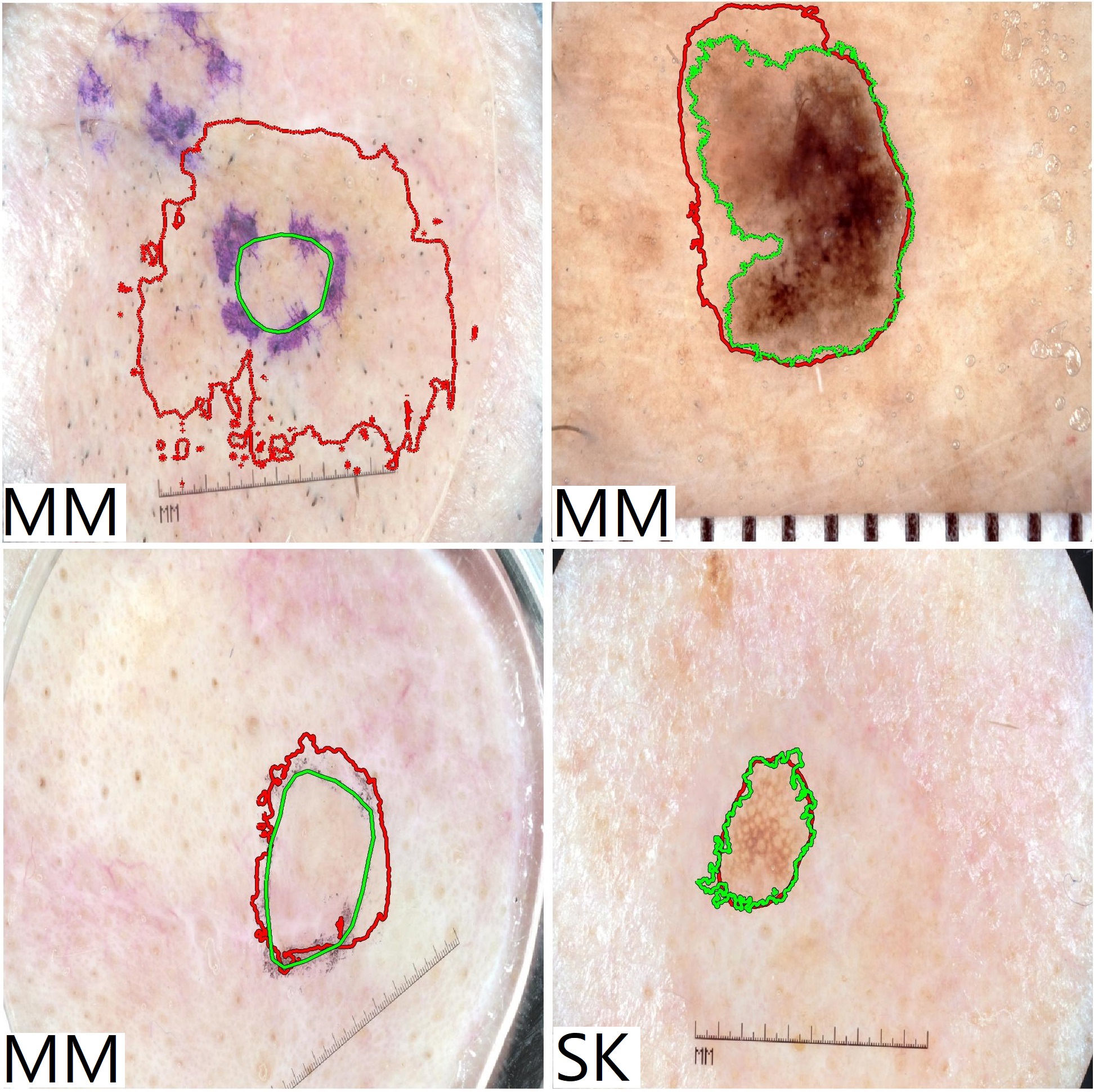}
	\end{center}
	\caption{Examples of test images for malignant melanoma vs. all classification task which were only classified correctly when the \textit{reference dataset} was used but incorrectly classified when the \textit{dilated cropped dataset} was utilised. The green and red annotations show the manual and automatically created lesion segmentations, respectively. The true skin lesion type is depicted in the left corner of each image. MM: Malignant Melanoma, SK: Seborrheic Keratosis. }
	\label{MM_ref}
\end{figure}

\begin{figure}[t]
	\begin{center}
		\includegraphics[width=0.8\linewidth]{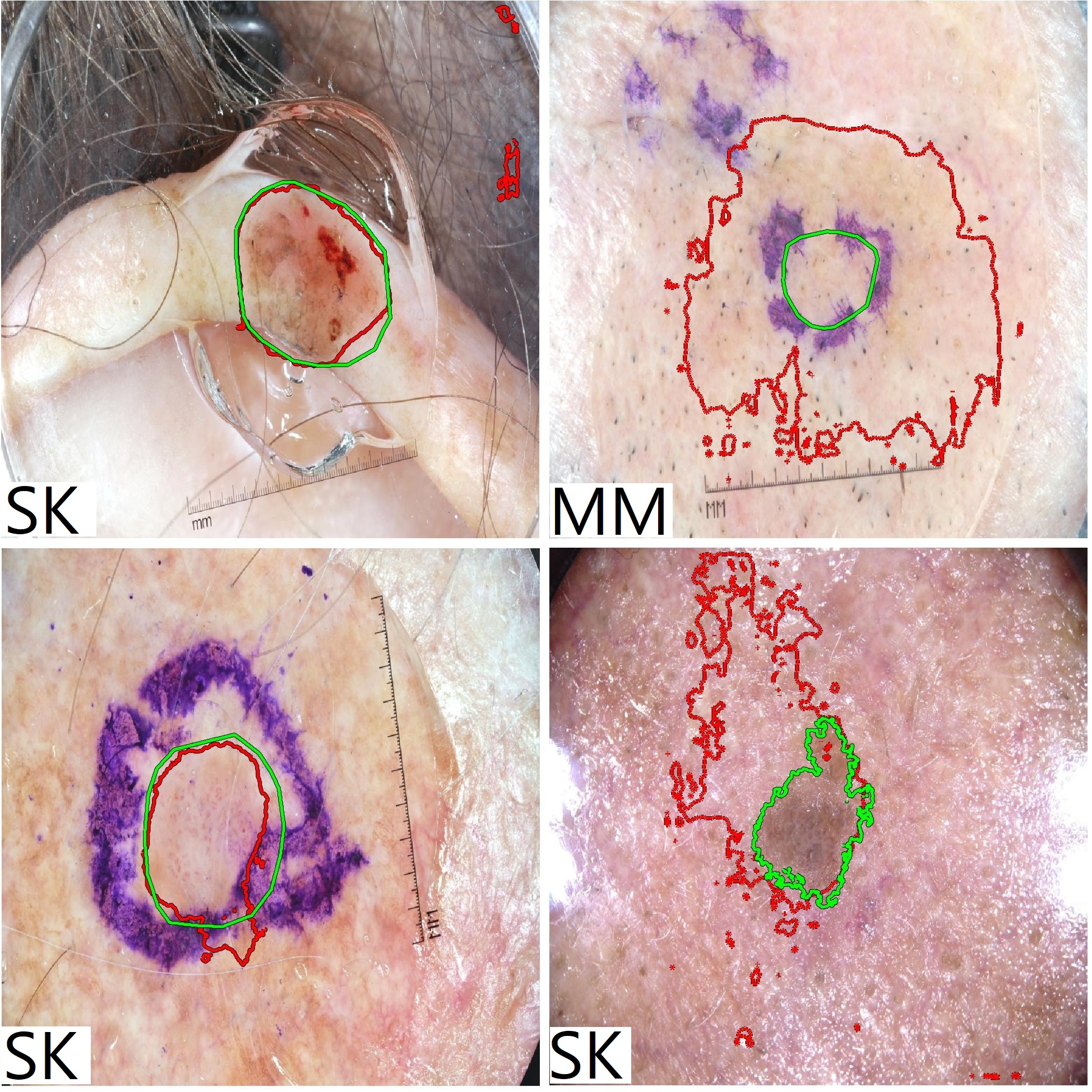}
	\end{center}
	\caption{Examples of test images for seborrheic keratosis vs. all classification task which were only classified correctly when the \textit{reference dataset} was used but incorrectly classified when the \textit{dilated cropped dataset} was utilised. The green and red annotations show the manual and automatically created lesion segmentations, respectively. The true skin lesion type is depicted in the left corner of each image. MM: Malignant Melanoma, SK: Seborrheic Keratosis. }
	\label{SK_ref}
\end{figure}

\begin{figure}[t]
	\begin{center}
		\includegraphics[width=0.8\linewidth]{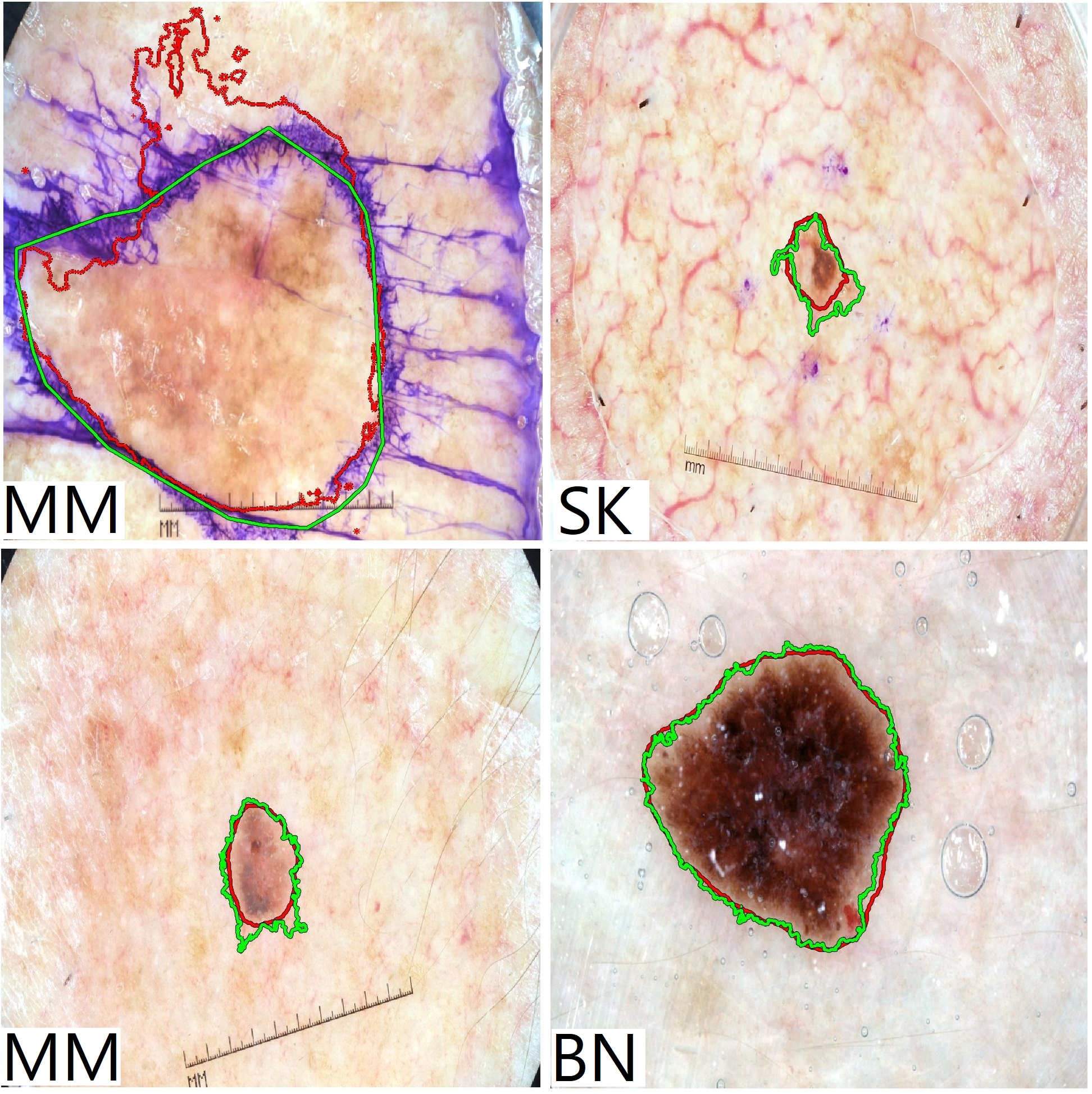}
	\end{center}
	\caption{Examples of test images for malignant melanoma vs. all classification task which were only classified correctly when the \textit{dilated cropped dataset} was used but incorrectly classified when the \textit{reference dataset} was utilised. The green and red annotations show the manual and automatically created lesion segmentations, respectively. The true skin lesion type is depicted in the left corner of each image. MM: Malignant Melanoma, BN: Benign Nevi, SK: Seborrheic Keratosis. }
	\label{MM_dilatecrop}
\end{figure}

\begin{figure}[t]
	\begin{center}
		\includegraphics[width=0.8\linewidth]{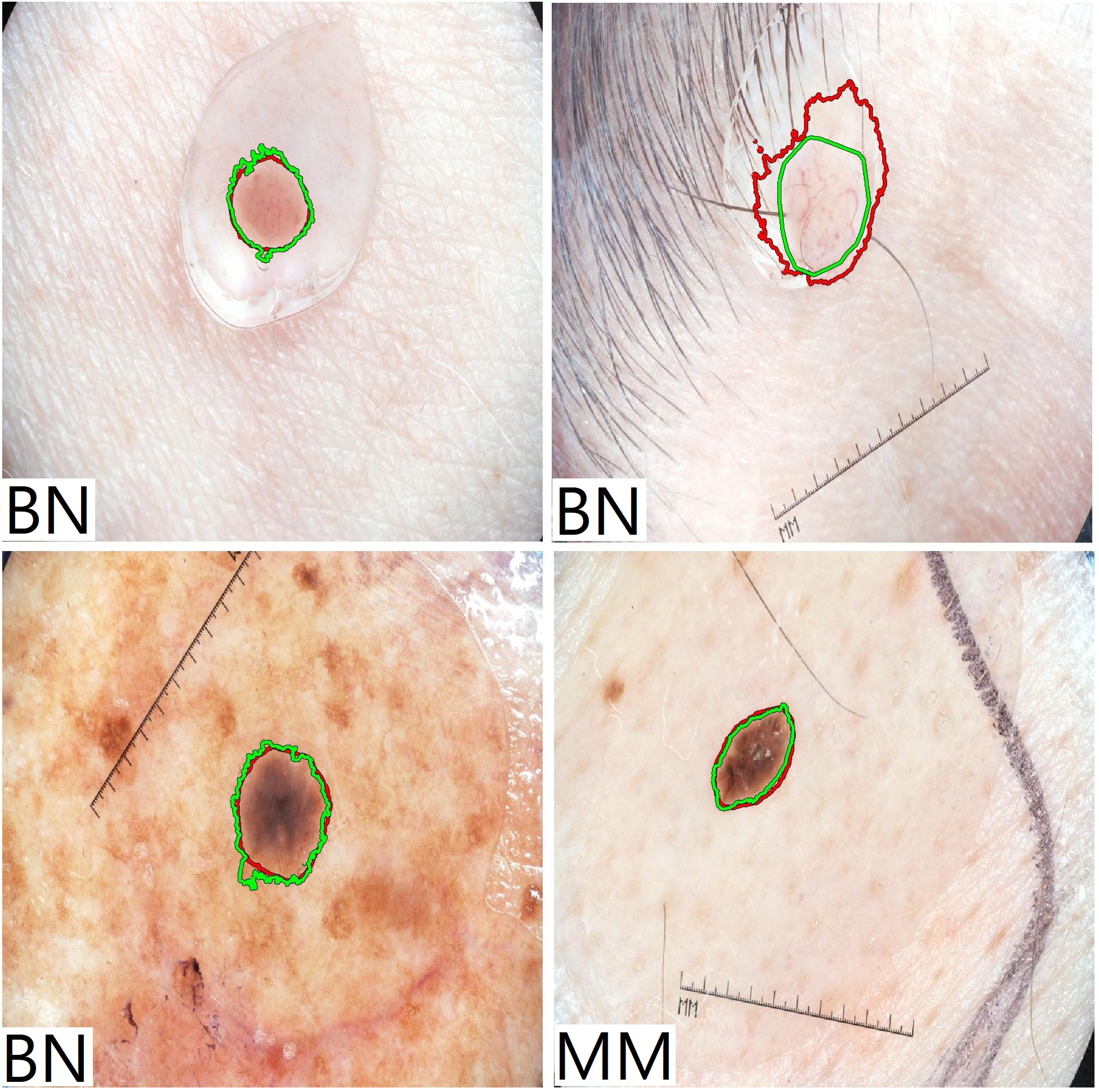}
	\end{center}
	\caption{Examples of test images for seborrheic keratosis vs. all classification task which were only classified correctly when the \textit{dilated cropped dataset} was used but incorrectly classified when the \textit{reference dataset} was utilised. The green and red annotations show the manual and automatically created lesion segmentations, respectively. The true skin lesion type is depicted in the left corner of each image. MM: Malignant Melanoma, BN: Benign Nevi. }
	\label{SK_dilatecrop}
\end{figure}

Fig.~\ref{MM_ref} and Fig.~\ref{SK_ref} show some examples of test images which are only classified correctly when the \textit{reference dataset} was used but incorrectly classified when the \textit{dilated cropped dataset} was utilised. On the other hand, Fig.~\ref{MM_dilatecrop} and Fig.~\ref{SK_dilatecrop} show some examples that are only classified correctly when \textit{dilated cropped dataset} was used but incorrectly classified when the \textit{reference dataset} was used. For visual comparison, we selected the \textit{dilated cropped dataset} as it has shown better overall classification performances compared to the other datasets.

\section{Discussion}
In this study, we explicitly investigated and explored the effect of using skin lesion segmentation on classification performance through a systematic process and using one of the well-known publicly available datasets (the ISIC 2017 challenge dataset). First, we developed a baseline classifier (without using any segmentation mask, any sophisticated ensemble strategy, and any external training dataset) that delivers a good classification performance comparable to the other state-of-the-art classification models. Then we conducted a comprehensive investigation to explore the effects of using segmentation masks on the skin lesion classification performance through two sets of experiments. We investigated the effects of both manually created segmentation masks (using the ground truth segmentation masks) and automatically created segmentation masks (using our developed \textit{SkinLinkNet} segmentation model) on the classification performance in 10 different scenarios. In addition to reporting the actual classification scores, we also performed statistical tests to evaluate whether the reported values were significantly different. All derived classification prediction vectors from different scenarios and also the automatically created segmentation masks are available from this Github repository\footnote{Upon acceptance of the paper, we will make the repository publicly available}:  \url{https://github.com/masih4/Skin-lesion-segmentation-effects-of-the-classification-perfromnce}  


The results reported  in Table.~\ref{classification_ref} show the comparative results of our single classification network and the top three performers of the ISIC 2017 challenge as well as four other state-of-the-art algorithms. While the overall classification performance of the developed model seemed slightly inferior compared to some other approaches, we could not find a significant difference in the performance. We were not able to perform statistical tests to compare the results with the first and third rank approaches as the prediction vectors of those two studies were not available. We found the classification prediction vector of the other studies either through the corresponding Github repositories or by asking from the paper's authors to share the prediction vectors.    

As explained in Section~\ref{pre-processing}, we chose the ISIC 2017 challenge dataset in this study as the ground truth for the test set of both segmentation and classification tasks were available. However, to further evaluate the performance of the baseline classification model, we used a similar approach and applied it on the ISIC 2018 challenge dataset. We used the training set of the ISIC 2018 dataset for training the baseline classifier and tested on the test set of the ISIC 2018 challenge. As the ISIC 2018 challenge images had a fixed image size of $450 \times 600$ pixels, we extracted random crops with a fixed size of $450 \times 450$  for the model training. Besides this difference, we used the identical training scheme as explained in Section~\ref{Classification_model}. Our single classification model, without using any external dataset and any ensembles, achieved an average recall score of 83.6\% on the test set of the ISIC 2018 challenge dataset.  Using external datasets for training and a straightforward fusion approach (as explained in our former study in~\cite{MAHBOD2020105475}), the performance was improved to 87.2\% 
which confirms a good classification performance of the developed method. Our single classification model and our fusion model currently rank 27th and 4th, respectively out of more than 200 participating teams and more than 11000 submissions in the ISIC 2018 online evaluation platform for skin lesion diagnosis. \footnote{\url{https://challenge2018.isic-archive.com/live-leaderboards/} Task 3: Lesion Diagnosis (Accessed on 2020-08-04)}.

 
As explained in Section~\ref{pre-processing}, in the first set of experiments, we used manual segmentation masks provided by the experts, which are the ideal segmentation masks. In the second set of experiments, we made use of automatically created segmentation masks generated by the developed \textit{SkinLinkNet} segmentation model. To make sure that our segmentation model could produce comparable results to the other state-of-the-art models, we compared the segmentation results with the top three performers of the ISIC 2017 challenge in the segmentation part as shown in Table.~\ref{segmentation_ref} (rows 1--3). As the results show, our utilised \textit{SkinLinkNet} segmentation model delivered slightly inferior segmentation performance compared to the top three performers of the ISIC 2017 challenge. However, after performing statistical tests, we observed no significant differences between the results. To extend the comparison, we also developed two additional segmentation models. One delivered slightly but still significantly inferior segmentation performance and the other one delivered slightly but still significantly superior segmentation performance as shown in Table.~\ref{segmentation_ref}.  It is worth mentioning that a direct comparison of the \textit{SkinLinkNet} and \textit{SkinFPN+} is not rational as the latter one was trained with extensive external data. As explained in Section~\ref{segmentation_model}, the main idea behind developing the additional segmentation models was to investigate if there is any significant differences between using manually created segmentation masks and automatically created segmentation masks (either through using \textit{SkinLinkNet}, \textit{SkinUNet}, or \textit{SkinFPN+}) on the classification performance (details in Table~\ref{dilated_crop} and Table~\ref{manualVSauto}).

The results in Table.~\ref{auc_compare} show the main finding of this study for both sets of experiments. Several interesting outcomes can be inferred from the results of this table. First, for most of the cases where segmentation masks were used, we could not observe significant differences in the performance in comparison to the reference results (i.e. using no segmentation masks). Second, using the segmentation masks did not significantly improve the MM classification performance in any scenario but significantly degraded it in some cases. Third, using the dilated cropping strategy, the SK classification performance was significantly improved which suggested that this method is the best way to use the segmentation masks in the classification workflow. Finally, for the scenarios where background information was completely removed or 4-channel input images were used, the overall classification performance was inferior compared to the reference results.

We performed similar comparisons between the results based on the other evaluation indexes as shown in Table.~\ref{acc_compare}. Here, only in one of the cases (using \textit{dilated cropped dataset}), the SK classification result was significantly improved. However, we used a threshold of 0.5 (as suggested in the ISIC 2017 challenge) to measure the accuracy, sensitivity, and specificity which may not be the optimal threshold. As only 600 images were used in the test phase, misclassifying only a few images with the utilised threshold could change the accuracy, sensitivity, and specificity drastically and hence the results may not show the effect of each utilised approach reliably. For a better evaluation, a bigger test set should be used in future studies to investigate the effect of using segmentation masks on the accuracy, sensitivity, and specificity. 

For visual evaluation, we show some examples in Fig.~\ref{MM_ref} and Fig.~\ref{SK_ref} that are only classified correctly when the \textit{reference dataset} was used but incorrectly classified when the \textit{dilated cropped dataset} was utilised. We selected  \textit{dilated cropped dataset} for comparison as it was shown to have the best classification performance among other scenarios (refer to Table~\ref{auc_compare} and Table~\ref{acc_compare}). The manual and automatic segmentation masks (using \textit{SkinLinkNet}) are shown as green and red overlaid contours on the raw images. 

On the other hand, Fig.~\ref{MM_dilatecrop} and Fig.~\ref{SK_dilatecrop} show the opposite cases where only the \textit{dilated cropped dataset} delivered the correct classification and the \textit{reference dataset} led to wrong classification. 

From these examples, we can assume that when the lesions are a very small part of the images, using segmentation masks for dilated cropping may lead to better classification performance in comparison to the baseline classifier. To investigate this effect quantitatively, we calculated the accuracy of the models using only the dermatoscopic images containing small skin lesions (i.e. ratio of the skin lesion to the entire image was less than 2\%). We observed a superior classification performance of the model that was trained with \textit{dilated cropped dataset} in comparison to the baseline classifier for both MM vs. all (94.3\% vs. 91.4\%) and SK vs. all (97.1\% vs. 94.3\%) classification tasks. However, further studies are needed to find out the underlying reasoning of the model prediction. A complete list of images that are classified only with one of the approaches for both classification problems (i.e. MM vs. all and SK vs. all) can be found in our Github repository. 

As apparent from the results in Table~\ref{auc_compare}, using the \textit{dilated cropped dataset} delivered the best overall classification performance. In Table~\ref{dilated_crop}, we explicitly compare the classification results from the \textit{reference dataset} with the \textit{dilated cropped dataset} when manually (through manual segmentation masks) or automatically (through using either \textit{SkinLinkNet}, \textit{SkinUNet}, or \textit{SkinFPN+}) created segmentation masked were used. The results in this table show that for all cases the overall classification performance of \textit{dilated cropped dataset} is better over the \textit{reference dataset}. The results also show a very competitive performance for the \textit{dilated cropped dataset} in different scenarios. As the results in Table~\ref{manualVSauto} suggest there are no statistical differences when manual segmentation masks or automatically created segmentation masks were used.  This means, upon a proper usage of segmentation masks, one can rely on the automated segmentation model to crop the images and then perform classification. Interestingly, it can be also inferred that there is no need to use the best possible automatic segmentation model to perform dilated cropping as it does not have a significant impact on the final classification performance. While the segmentation performance of the \textit{SegUNet} is statistically inferior compared to \textit{SkinLinkNet} and the segmentation performance of the \textit{SkinLinkNet} is statistically inferior compared to \textit{SkinFPN+}, all delivered comparable classification performance to the manual segmentation masks. 


\section{Conclusion}
In this paper, we have explored the effects of using skin lesion segmentation masks on the performance of dermatoscopic image classification. Using a baseline classification network and using manually or automatically created segmentation masks in different scenarios, we observed several interesting outcomes. Our results suggest that using segmentation masks in a proper way can significantly improve the overall classification performance. However, using the masks in an inappropriate manner by removing all background information can significantly degrade the classification results. Moreover, we show that by proper exploitation of segmentation masks, there is no significant difference in the classification performance when manually or automatically created segmentation masks are used. 

\section*{Conflict of interest statement}
There are no conflicts of interest to disclose for publication of this paper.

\section*{Acknowledgment}
This work was supported by the EU Horizon 2020 CaSR Biomedicine project, No. 675228 and the Austrian Research Promotion Agency (FFG), No. 872636. The authors would like to thank the TissueGnostics Research and
Development team\footnote{\url{https://www.tissuegnostics.com}} for valuable suggestions. Moreover, we thank NVIDIA corporation for their generous GPU donation.

\bibliography{invres}

\end{document}